\documentclass{article}
\usepackage{graphicx}
\usepackage{amsmath}
\usepackage{url}
\usepackage{hyperref}
\usepackage{booktabs}
\usepackage{pgfplotstable}

\title{From Predictive Importance to Causality: Which Machine Learning Model Reflects Reality?}

\author{
    Muhammad Arbab Arshad \and Pallavi Kandanur \and Saurabh Sonawani \\
    Department of Computer Science, Iowa State University \\
    50014, United States \\
    \texttt{\{arbab, Pallavik, sonawani\}@iastate.edu}
    \and
    Laiba Batool \\
    National University of Computer \& Emerging Sciences \\
    Islamabad 44000, Pakistan \\
    \texttt{laiba.batool.1426@gmail.com}
    \and
    Muhammad Umar Habib \\
    Lahore University of Management Sciences \\
    Sector U, DHA, Lahore, Pakistan \\
    \texttt{muhammadumarhabib@gmail.com}
}

\begin{document}

\maketitle

\begin{abstract}
This study analyzes the Ames Housing Dataset using CatBoost and LightGBM models to explore feature importance and causal relationships in housing price prediction. We examine the correlation between SHAP values and EconML predictions, achieving high accuracy in price forecasting. Our analysis reveals a moderate Spearman rank correlation of 0.48 between SHAP-based feature importance and causally significant features, highlighting the complexity of aligning predictive modeling with causal understanding in housing market analysis. Through extensive causal analysis, including heterogeneity exploration and policy tree interpretation, we provide insights into how specific features like porches impact housing prices across various scenarios. This work underscores the need for integrated approaches that combine predictive power with causal insights in real estate valuation, offering valuable guidance for stakeholders in the industry.
\end{abstract}

\section{Introduction}
The field of predictive modeling has seen remarkable advancements, where accurate predictions are the cornerstone of informed decision-making. However, alongside prediction lies a parallel pursuit of understanding the causal structure that underlies the data. This dual objective has become increasingly vital, especially in domains where uncovering the true drivers of outcomes is of paramount importance.

The Ames Housing Dataset, a meticulously curated compilation of nearly every conceivable facet of residential homes in Ames, Iowa, serves as a compelling platform for this endeavor. Covering transactions from 2006 to 2010, this dataset encapsulates a wealth of information, from zoning classifications to the very contour of the land. Its 79 diverse features promise insights that transcend mere correlations, offering the potential to discern the features that truly influence house prices.

The significance of this analysis extends beyond the realm of predictive modeling. Understanding the causal factors driving housing prices in Ames not only provides actionable insights for buyers, sellers, and policymakers in the local real estate market but also contributes to the broader discourse on real estate economics. It offers a lens into the intricate interplay of variables that dictate property values, which in turn can inform urban planning strategies and investment decisions.

In the following sections, we outline our approach, which combines advanced modeling techniques like LightGBM and CatBoost, with cutting-edge interpretability tools such as SHAP values and Econ ML. This approach, while emphasizing predictive power, also unlocks the potential to uncover the causal narratives embedded within the dataset. Through this project, we aim not only to produce accurate predictions but also to shed light on the underlying structure and intricacies of the housing market in Ames.
\subsection{Dataset Overview}
The Ames Housing Dataset stands as a comprehensive repository of housing information, meticulously compiled to encompass various facets of residential properties in Ames, Iowa. Covering transactions spanning the years 2006 to 2010, this dataset encapsulates a rich array of features, offering a nuanced perspective on the determinants of housing prices.

With approximately 2900 observations, the dataset strikes a balance between comprehensiveness and manageability, rendering it an ideal subject for in-depth analysis and modeling. Encompassing a total of 79 features, including both numerical and categorical variables, it paints a detailed picture of each property's characteristics. These features range from structural attributes like square footage and number of bedrooms, to contextual elements such as zoning classifications and proximity to various amenities.

At the heart of this dataset lies the sale price of each property, serving as the primary target variable for predictive modeling. This critical piece of information anchors our analysis, allowing us to unravel the factors that significantly influence real estate valuations.

While known for its relative cleanliness, it's essential to acknowledge that, like any real-world dataset, the presence of outliers may necessitate careful consideration during the analysis. Thus, a meticulous approach to data preprocessing and exploratory data analysis will be pivotal in ensuring the integrity and reliability of our findings.

In the subsequent sections, we delve into our methodology, outlining the techniques and models that will be employed to extract valuable insights from this robust dataset.
\subsection{Motivation and Scope}
In the ever-evolving landscape of predictive modeling, understanding the underlying causal structure of a dataset is equally as vital as producing accurate predictions. The Ames Housing dataset, a comprehensive compilation of almost every facet of residential homes in Ames, Iowa, poses a unique opportunity to probe deeper into this conjunction of predictive power and causality.

Our primary goal is to employ two distinct models: CatBoost (Categorical Boosting) and LightGBM (Light Gradient Boosting Machine). By training each on the Ames dataset and subsequently extracting the top $k$ influential features using SHAP values, we aim to uncover which of these models' predictive features align most closely with genuine causal factors. This intersection of SHAP-driven predictive importance with true causal significance could reveal insights into which model not only predicts well but also serendipitously unravels the true causal narratives embedded within the dataset.

Moreover, the investigation will not just stop at individual feature importance. By aggregating the top features from each model, we'll create a feature superset that will undergo advanced causal inference. This approach guarantees that we aren't confined by the predictive prowess of a singular model, but we amalgamate the collective insights of all two. The finale of this exploration will be a rank comparison, juxtaposing the order of features based on global causal importance against their SHAP rankings.

In essence, our journey is not merely to find a model that predicts house prices accurately but to unveil which model, in its predictions, also serendipitously unravels the true causal narratives embedded within the dataset.

The analysis will be primarily focused on the dataset itself, with limited consideration of external factors that may influence housing prices, such as macroeconomic indicators or policy changes. While outliers and missing values will be addressed as part of standard data cleaning procedures, extensive data imputation or complex outlier handling falls outside the scope.

Given the established nature of the dataset, this project is not focused on the creation of a predictive model for practical deployment, but rather on extracting meaningful insights and understanding the underlying causal structure within the Ames Housing Dataset.
\subsection{Objective}

\subsection{Primary Objective}
The primary objective of this project is to conduct a comprehensive analysis of the Ames Housing Dataset, focusing on three key areas:

\begin{enumerate}
    \item Determining Feature Importance: Using CatBoost and LightGBM to identify features significantly influencing housing prices.
    
    \item Uncovering Causal Relationships: Applying causal inference to discern genuine causal factors, providing deeper insights into Ames' housing market dynamics.
    
    \item Exploring Correlation between SHAP Values and Econ ML Predictions: Investigating the relationship between SHAP values and predictions derived from economic machine learning models.
\end{enumerate}

\subsection{Secondary Objectives}
In addition to the primary objectives outlined above, this project aims to achieve the following secondary objectives:

\begin{enumerate}
    \item Evaluate Model Performance: Assessing the effectiveness of the chosen models (CatBoost, LightGBM) in accurately predicting housing prices within the Ames dataset.
    
    \item Compare Feature Importance Rankings: Comparing the feature importance rankings obtained from different models to identify consistent influential factors.
\end{enumerate}

These objectives collectively form the foundation of this project, guiding the analysis and providing a framework for evaluating the results.

\section{Related Work}

\subsection{Predictive Modeling in Real Estate}

Predictive modeling in real estate has gained substantial traction in recent years. Researchers and practitioners alike recognize its potential for informing critical decisions related to property transactions, investment strategies, and urban planning \cite{de2011ames}. By harnessing the power of machine learning, analysts can distill valuable insights from vast datasets, offering a data-driven approach to understanding complex real estate dynamics.

\subsection{Interpretable Machine Learning}

The adoption of interpretable machine learning techniques has emerged as a crucial aspect of model deployment in real-world applications. Methods like SHapley Additive exPlanations (SHAP) values provide a clear and intuitive way to understand the impact of individual features on model predictions \cite{NIPS2017_7062}. By unraveling the 'black box' of complex models, interpretability tools enhance transparency and build trust in the predictive process.

\subsection{Causal Inference in Real Estate}

Understanding causal relationships in real estate is paramount for making informed policy decisions and investment strategies. Recent advances in causal inference, as exemplified by libraries like EconML, empower analysts to disentangle correlation from causation, shedding light on the true drivers of property values \cite{econml}. This approach goes beyond prediction, providing a deeper understanding of the underlying mechanisms shaping the real estate market.

\subsection{Advanced Techniques: CatBoost and LightGBM}

In the realm of predictive modeling, CatBoost and LightGBM stand out as state-of-the-art techniques. CatBoost's innate capability to handle categorical variables and LightGBM's leaf-wise tree growth strategy render them powerful tools for accurate predictions \cite{LGBM}. Leveraging the strengths of both these models holds great promise for unraveling the intricate web of factors influencing housing prices.

The convergence of advanced predictive modeling techniques \cite{jha2023housing}, interpretable machine learning, causal inference, and ensemble learning presents a compelling opportunity to dissect the complex interplay of factors driving housing prices in Ames, Iowa. By integrating these methodologies, this project aims not only to produce accurate predictions but also to uncover the true causal narratives embedded within the dataset. Recent advancements in large language models may also offer new perspectives on data analysis and interpretation in various domains \cite{shahriar2024putting}.

``Causal Interpretation for Ames Housing Price'' is an important reference for its in-depth exploration of causal relationships in real estate data. Additionally, the Ames Housing Dataset, a meticulously curated compilation of residential property information, serves as a foundational resource for this analysis \cite{econml2023}.

``A Unified Approach to Interpreting Model Predictions'' provides valuable insights into the importance of interpretability in machine learning, a principle that underpins the methodology of this project \cite{NIPS2017_7062}.

In the next sections, we detail our data exploration, preprocessing techniques, and the specific methodologies employed, including the application of CatBoost and LightGBM, as well as the implementation of EconML for causal inference.

\section{Methodology:}

\subsection{Data Preparation}

To achieve our goal,f the Ames Housing Dataset, a comprehensive collection of data from Kaggle.. This dataset has 2,930 distinct observations, each representing a unique residential property in Ames, Iowa. This dataset has 79 different features that describe a residential property. The target variable here is the sale price. The details are given in \ref{data1} and \ref{data2}

The dataset comprises a variety of categorical and numerical values. We will detail how we processed and handled these values in the subsequent section.

\section{Data Preprocessing}

The dataset underwent a rigorous preprocessing regime informed by a previous study \cite{econml2023}. The overarching motivation for this preprocessing was to delve into the correlation between SHAP values, a prominent interpretability mechanism, and predictions derived from economic machine learning models. The primary focus was on "Exploring Correlation between SHAP Values and Econ ML Predictions: Investigating the relationship between SHAP values and predictions originating from economic machine learning models."

Initially, we augmented the dataset with the \texttt{SalePrice} as a column, representing our target variable, and set the dataset's index to the \texttt{Id} column for better accessibility. To preserve the integrity of our data, entries with missing data in the \texttt{MasVnrType} and \texttt{Electrical} columns were eliminated.

Subsequent to these steps, we derived several columns from the extant dataset, aiming to capture nuanced insights:
\begin{itemize}
    \item \texttt{AgeAtSale}: Calculated by finding the difference between the year a property was sold (\texttt{YrSold}) and its construction year (\texttt{YearBuilt}).
    \item \texttt{YearsSinceRemodel}: Represents the years elapsed since the last remodeling, ensuring no negative values.
    \item \texttt{HasDeck}: A binary marker signifying the presence of a deck in the property.
    \item \texttt{HasPorch}: Flags the presence of any form of porch in the dwelling.
    \item \texttt{HasFireplace}: A binary indicator depicting the existence of a fireplace.
    \item \texttt{HasFence}: Indicates the property's fencing status.
\end{itemize}

To refine our dataset further and obviate potential multicollinearity, we discarded a swath of columns, some of which pertained to various measurements of area and temporal features like \texttt{GarageYrBlt}, \texttt{YearBuilt}, and \texttt{YrSold}. These columns were either superfluous due to our newly derived metrics or were judged as tangential to our primary analysis.

In terms of data imputation, the missing values in the \texttt{LotFrontage} column were supplanted with zeros, operating under the assumption of no adjoining street. Other lacunae across the dataset were filled with 'NA', symbolizing the absence of a particular feature. The categorical variables were handled inherently by our selected models, eliminating the necessity for explicit one-hot encoding or preprocessing. As exemplified in the code, the categorical columns were converted to the category data type, which is favored by models like LightGBM.

\subsection{Model Definitions}
\subsubsection{LightGBM}
Light Gradient Boosting Machine is a powerful gradient boosting framework, based on decision trees. \\
Gradient Boosting is a technique that combines multiple weak models to generate a powerful model. This final model gives accurate predictions by iteratively correcting the errors generated by the weaker models.After each model (tree) is trained, the residuals or errors are analogous to the gradient of a loss function with respect to the model's predictions. In gradient boosting, the predictions are adjusted in the direction that decreases the loss thus using the concept of gradient descent.\\
\begin{figure}[h!]  
  \includegraphics[width=\linewidth]{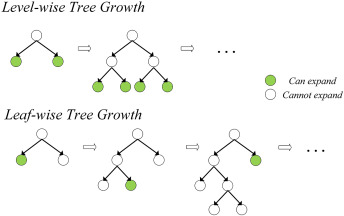}
  \caption{\label{fig:workflow}
     Internal of LightGBM and CatBoost
    \cite{Yao2022}
  }
\end{figure}

LightGBM uses the leaf -wise splitting technique where the tree grows by splitting the node that results in the maximum reduction of the loss function, regardless of its depth in the tree, which allows it to to converge faster and often achieve better performance ~\ref{fig:workflow}. For handling high dimensional data it uses the technique of Exclusive Feature Bundling (EFB) , in which two columns whose values are mutually exclusive , (can never be the same at once) are bundled together into a single feature. This reduces dimensionality and improves accuracy.
To speed up the training process , LightGBM uses GOSS(Gradient-based One-Side Sampling).In this technique, the data points with larger gradients are considered more important for the training process as they are misclassified. GOSS retains all the larger gradient data points and then from the lower value instead of using all, it selects a fraction. This improves speed and efficiency and reduces memory consumption.\\

In order to optimize model performance, we employed a grid search methodology using the \texttt{GridSearchCV} module from Scikit-Learn library.  The fine-tuning was primarily focused on four parameters \texttt{learning\_rate},  \texttt{max\_depth}, \texttt{num\_leaves}, \texttt{min\_child\_samples}.( The parameters here are self explanatory)
The rest of the hyperparameters were used in their default settings.
The range for  \texttt{learning\_rate} was [0.1, 0.05, 0.01] while the model was evaluated on \texttt{max\_depth} for [3, 5, 10]. We dynamically computed the range for \texttt{num\_leaves} parameter based on the given \texttt{max\_depth}, setting it to be $2^depth$ depth for each depth in the \texttt{max\_depth} list.For the  \texttt{min\_child\_samples} the range was [20, 30, 40] \\

Upon completion of the search, we identified the optimal value combination for our model to be: \texttt{learning\_rate} of 0.1, \texttt{max\_depth} of 10,  \texttt{num\_leaves} of 8 and  \texttt{min\_child\_samples} to be 20.

\subsubsection{CatBoost}
 CatBoost is a high-performance, open-source gradient boosting library developed by Yandex, a Russian multinational corporation specializing in Internet-related products and services. CatBoost stands for "Categorical Boosting" . 
 Similar to LightGBM , CatBoost also uses gradient boosting technique.  However it employs the level wise growth architecture ~\ref{fig:workflow} which means that all nodes at the current depth (level) are expanded before moving to the next level. Every leaf node is at the same depth which helps prevent overfitting.\\
It has the capability of handling the categorical data directly without specifying the datatype. It uses the technique of Ordered boosting when encoding a categorical variable for a specific row by only using information from rows before it so as to avoid target leakage. 

To get optimal model performance,  we used the same approach as in LightGBM - \texttt{GridSearchCV} . The hyperparameters which we attempted to tune were \texttt{learning\_rate},  \texttt{depth}, \texttt{l2\_leaf\_reg}, \texttt{border\_count}.Here the  \texttt{l2\_leaf\_reg} is the Coefficient at the L2 regularization term of the cost function and \texttt{border\_count} defines number of bins in the numerical features.The parameters input to the grid search were :  \texttt{learning\_rate} was [0.1, 0.05, 0.01] , \texttt{depth} for [3, 5, 10],  \texttt{l2\_leaf\_reg} values [1, 5, 10] and \texttt{border\_count} to be  [32, 128, 255].\\

The model performed well when the hyperparameters had the values : 

\subsection{Model Interpretability}

\subsubsection{SHAP Values}
SHAP(SHapley Additive exPlanations) values are a way to express how much each feature is affecting the model prediction or how much each feature contributes to the model prediction. These values are used to interpret machine learning models and improve explainability. \\
These values are computed based on the concept of SHAPLEY values in game theory.
First subsets are created excluding the feature for which we want to calculate the SHAP values. Then for each subset, model prediction is calculated . Now, model predictions is calculated again for that subset but with the feature in question. Now, the difference in the model's prediction before and after adding the given feature is measured and the average of these marginal contributions of that feature across all possible permutations of features is taken .
Using these techniques, shap library provides interpretable feature importances that have a strong theoretical foundation and offer insights into how individual features impact predictions.\\

We have used this library to calculate and visualize the important features for our model.

\subsection{Causal Inference and Effect Estimation }
\subsubsection{EconML}
EconML is a Python library developed by Microsoft Research that leverages recent advances in machine learning to estimate causal effects. 
For our analysis, it is important to understand the causal relationships between the target variable and the individual features. This will help in understanding how the housing prices change if a certain feature is changed, For instance, does adding a porch have a different impact in a residential versus a commercial zone?
In our project, we've applied the CausalAnalysis module from EconML. This approach allows us to:\\
Isolate True Causal Drivers: By distinguishing between mere correlation and causation, we can pinpoint which housing features genuinely drive price changes. \\
Enhance Predictive Models: By ensuring our model is not basing predictions on correlations we can be more confident in its predictions and recommendations.

\subsubsection{Average Treatment Effect}
We are using the 'global causal effect' method from the 'CausalAnalysis' module in the EconML.This method provides an ATE for each feature in the dataset, offering a broad view of the potential causal relationships within the data.
The ATE essentially provides an understanding of the expected change in the outcome for a one-unit change in the feature, assuming all other factors remain constant.For instance, if we have a feature  "HasFireplace", the ATE would give us:
The expected difference in house prices between homes with a fireplace and those without, considering all other features remain unchanged.
This will be done for all the features one by one. Since there are multiple features and we want to understand the effect one by one , we are using 'global causal effect'.

In essence, while traditional analyses might tell you what is happening, the global causal effect helps explain why it's happening, which is invaluable in many practical scenarios.

\subsection{Evaluation Techniques}
\subsubsection{Spearman's Rank Correlation}

We aim to compute the Spearman's rank correlation coefficient between a list of causally significant features, denoted as \( C \), and a list of features determined via SHAP analysis for a specific model, denoted as \( F_m \).

\textbf{Definitions and Notations:}
\begin{itemize}
    \item \( C \): The list of causally significant features obtained from a specific causal analysis. This list is sorted in a particular order and represents a subset of all considered features.
    \item \( F_m \): The list of features sorted based on their importance for model \( m \) as determined by a SHAP analysis.
    \item \( C_m \): Represents the intersection of features between \( C \) and \( F_m \).
\end{itemize}

The common features between the lists \( C \) and \( F_m \) are:
\[ C_m = \{ f \mid f \in F_m \text{ and } f \in C \} \]

For each feature \( f \) in \( C_m \), its rank is determined based on its position in \( C \) and \( F_m \):
\begin{align*}
    r_C(f) & = \text{position of } f \text{ in } C \\
    r_{F_m}(f) & = \text{position of } f \text{ in } F_m \\
\end{align*}
From this, we derive two ranked lists: 
\begin{itemize}
    \item \( R_C \): Ranks of common features based on their order in \( C \).
    \item \( R_{F_m} \): Ranks of common features based on their order in \( F_m \).
\end{itemize}

The Spearman's rank correlation coefficient, \( \rho \), for the ranked lists \( R_C \) and \( R_{F_m} \) is computed using:
\[ \rho = 1 - \frac{6 \sum_{i=1}^{n} d_i^2}{n(n^2 - 1)} \]
Where:
\begin{itemize}
    \item \( d_i \) represents the difference between the ranks of the \( i^{th} \) feature in the two lists.
    \item \( n \) denotes the number of features in \( C_m \).
\end{itemize}

\textbf{Interpretation:}
The Spearman's rank correlation coefficient, \( \rho \), yields a value between -1 and 1. A value of \( \rho = 1 \) indicates a perfect positive correlation, meaning that the order of the rankings in \( R_C \) and \( R_{F_m} \) are identical. Conversely, a value of \( \rho = -1 \) indicates a perfect negative correlation, signifying that the order of rankings in \( R_C \) is the exact reverse of \( R_{F_m} \). A value of \( \rho = 0 \) suggests no correlation between the rankings of the two lists.

\section{Experiments}

\subsection{Model Training}
After the steps of preprocessing and feature engineering, the data was segregated into a training and testing set, using an 80-20 split. A comprehensive grid search was executed for each model over a multitude of hyperparameters, as elucidated in the previous section (refer Table~\ref{tab:hyperparameters}). During training, a 5-fold cross-validation was performed to ensure the robustness of the models.

For LightGBM, the model achieved a score of \textbf{86.48} on the test set. In contrast, CatBoost showcased a score of \textbf{89.91} on the test set.

\begin{table}[ht]
\centering
\caption{Hyperparameters for LightGBM and CatBoost models}
\label{tab:hyperparameters}
\begin{tabular}{lcc}
\toprule
Parameter & LightGBM & CatBoost \\
\midrule
Learning Rate & 0.1, 0.05, 0.01 & 0.1, 0.05, 0.01 \\
Max Depth & 3, 5, 10 & 3, 5, 10 \\
Num Leaves & $2^{\text{Max Depth}}$ & N/A \\
Min Child Samples & 20, 30, 40 & N/A \\
L2 Leaf Reg & N/A & 1, 5, 10 \\
Border Count & N/A & 32, 128, 255 \\
\bottomrule
\end{tabular}
\end{table}

\subsection{Feature Importance Analysis}
\begin{figure*}[htbp]
    \centering
    \includegraphics[width=0.9\textwidth]{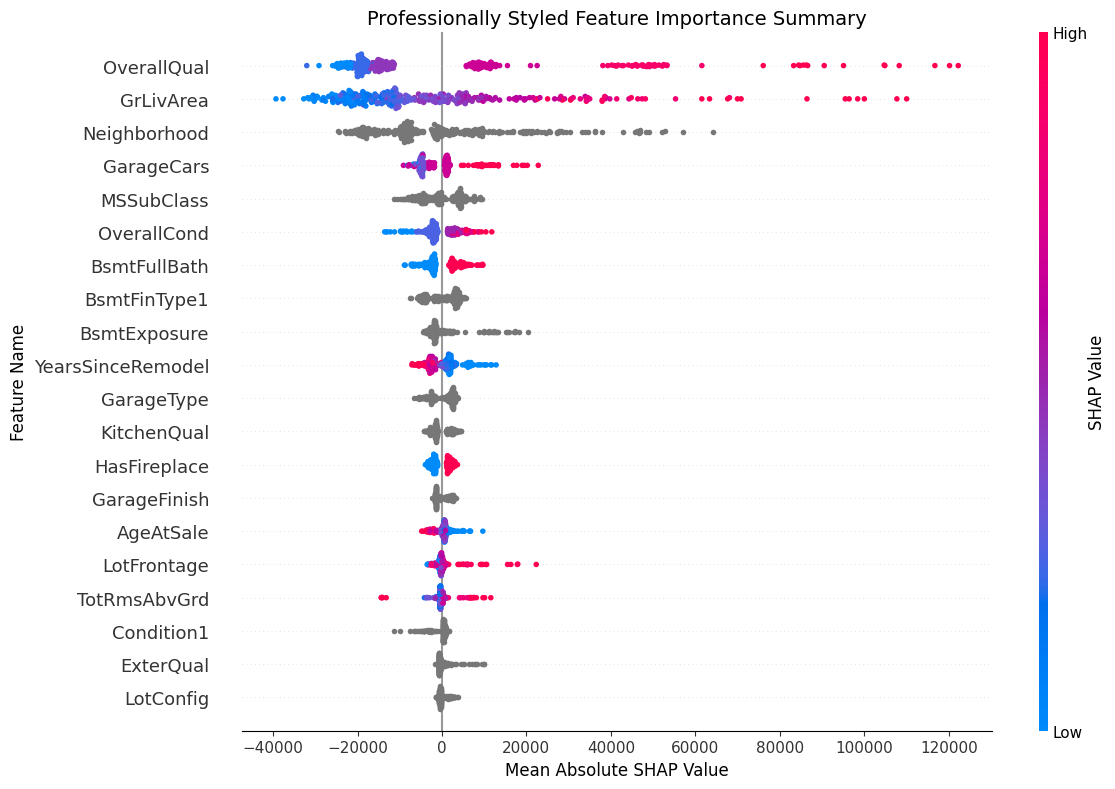}
    \caption{SHAP Values Visualization for LightGBM}
    \label{fig:shap_lightgbm}
    
    \vspace{1cm}
    
    \includegraphics[width=0.9\textwidth]{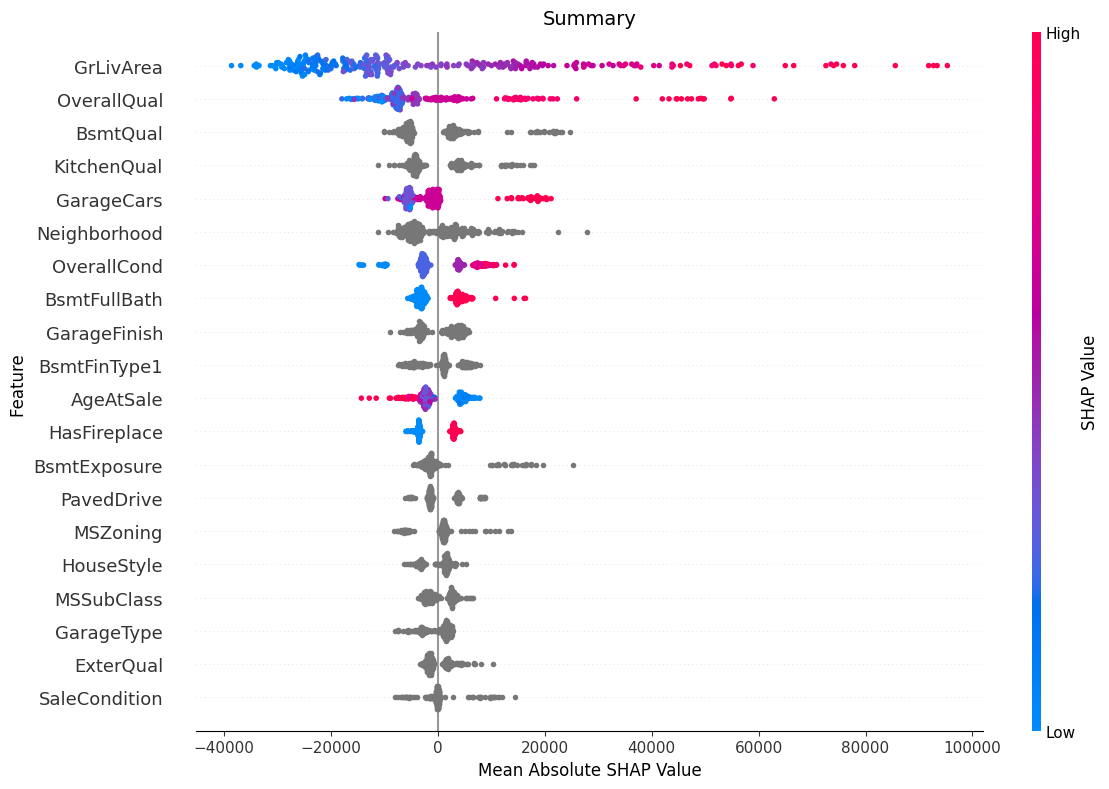}
    \caption{SHAP Values Visualization for CatBoost}
    \label{fig:shap_catboost}
\end{figure*}

To comprehend the model's decisions and dissect the contributions of each feature to the predictions, we employed the SHAP (SHapley Additive exPlanations) framework. Particularly, we used the \texttt{TreeExplainer} from the SHAP library, optimized for tree-based models, leveraging the Tree SHAP algorithms. As seen in Figure \ref{fig:shap_lightgbm}, and Figure \ref{fig:shap_catboost}

For our model, the TreeExplainer was initialized with several parameters:
\begin{itemize}
    \item \textbf{Model}: The trained model for which we want to determine the SHAP values.
    \item \textbf{Feature Perturbation}: Set to `tree\_path\_dependent`, this parameter determines how to address correlated or dependent features when estimating SHAP values. The tree path dependent method follows the structure of the trees, utilizing the count of training samples traversing each tree path as our background distribution. This approach abstains from requiring an explicit background dataset, simplifying the process.
    \item \textbf{Feature Names}: The names of the features in the training dataset were also provided for better interpretability.
\end{itemize}

Post initialization, SHAP values for the test set were computed. It is imperative to note that the SHAP framework, through \texttt{TreeExplainer}, allows for various output modes such as raw output, probability space, or even the log loss of the model. In our context, we focused on the raw output, which, depending on the model, can differ. For instance, in regression models, it typically represents the standard prediction, while for binary classification in frameworks like XGBoost, it reflects the log odds ratio.

The utility of SHAP values is manifold. They not only present a granular breakdown of feature influence but also offer insights that are rooted in game theory, ensuring equitable allocation of contributions. By adopting this approach, we aim to bridge the gap between model transparency and predictive performance, paving the way for more interpretable machine learning in housing price prediction.

\subsection{Causal Inference Examination}

Table \ref{tab:insignificant} presents features deemed unsuitable for causal analysis, along with their respective justifications. For our analysis, we utilized the \texttt{CausalAnalysis} tool from the \texttt{econml.solutions} package to explore causal relationships within the dataset's features. This tool facilitates identifying the causal effects of particular features on the outcome, while also considering heterogeneity across different attributes. Our training incorporated an extensive feature set, which included notable attributes like the presence of fireplaces, porches, and decks. Benefiting from parallel processing, our analytical approach was not only rigorous but also computationally efficient. After setting up the necessary parameters, we trained the model on our dataset. Results for few of the most casually significant features are given in \ref{tab: causal_significance}

\begin{table}[ht]
\centering
\caption{Top Casually Significant Features}
\label{tab: causal_significance}
\begin{tabular}{|l|l|l|r|}
\hline
\textbf{Feature} & \textbf{Feature Value} & \textbf{P-value} \\
\hline
BldgType & Twnhsv1Fam  & 0.000000e+00 \\
 & Duplexv1Fam  & 1.737315e-87 \\
OverallQual & num & 2.164378e-08 \\
GarageFinish & RFinFin  & 2.470622e-04 \\
BsmtExposure & GdvAv  & 4.727197e-04 \\
HasFireplace & 1v0 & 6.727178e-04 \\
GarageFinish & UnfFin & 8.408833e-04 \\
KitchenAbvGr & num 2 & 3.082170e-03 \\
GarageCars & num & 4.782600e-03 \\
\hline
\end{tabular}
\end{table}

\subsection{Relation between SHAP and Economic ML Predictions}
For LightGBM, the spearman metric value between the high ranked shap features and the casually significant features was calculated to be 0.35. On the other hand, CatBoost showcased a slightly higher value of 0.48.

This suggests that while both models agree to some extent on the importance of common features, the degree of agreement varies. CatBoost's higher correlation might hint at its better capability to align its feature importance with causal factors. It's also noteworthy to mention that CatBoost achieved a higher accuracy on the test data, scoring 89.91 compared to LightGBM's 86.48. This might be indicative of CatBoost's superior ability to generalize and potentially its better alignment with causally significant features, contributing to its enhanced performance on the test set.
\section{Extensive Causal Analysis}
\subsection{ Exploring Heterogeneity: The Impact of Porches on Housing Prices}
This section zeroes in on the 'HasPorch' feature in the Ames Housing Dataset, dissecting how porches affect property prices differently across various scenarios. Our goal is to understand these effects in detail, considering factors like property age, zoning, and additional features.\\
Rather than just identifying trends, we aim to unravel specific patterns in the causal relationship between porches and housing prices. This thorough analysis provides crucial insights for those in real estate, offering a nuanced understanding of how porches contribute to the pricing dynamics of residential properties.\\
To accomplish this, we use a tree-based method, visualizing a tree specifically for the 'HasPorch' feature. This allows us to pinpoint groups where porch impact on housing prices may vary. By interpreting the tree's nodes and branches, we uncover situations where having a porch significantly influences property values.
\begin{figure}[h!]  
  \includegraphics[width=\linewidth]{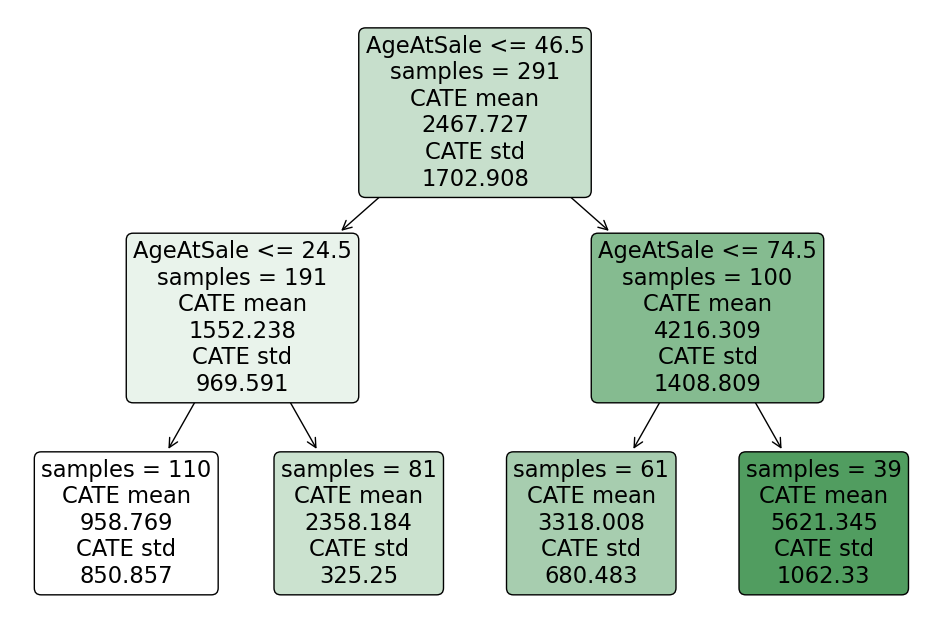}
  \caption{\label{fig:workflow}
     How do different type of houses respond differently to having a Porch?}
\end{figure}
The root node, with an age at sale less than or equal to 46.5 years, shows a moderate Conditional Average Treatment Effect (CATE) mean of 2467.727. This means that, on average, having a porch in properties of this age bracket contributes significantly to a house's value.However, delving deeper into the tree, we notice intriguing nuances. The left child node, representing properties with an age at sale less than or equal to 24.5 years, has a slightly lower CATE mean of 1552.238. This implies that, within this younger subset, the impact of having a porch is slightly diminished compared to the broader category.\\
On the other hand, the right child node, encompassing properties with an age at sale less than or equal to 74.5 years, exhibits a substantially higher CATE mean of 4216.309. This suggests that, for older properties in this range, the presence of a porch has a more pronounced positive effect on housing prices.

\subsection{Policy Tree Interpretation: Guiding Strategic Decisions}
As we shift our focus to the policy tree for the 'HasPorch' feature, our aim is to provide actionable insights for stakeholders in the real estate industry. Unlike heterogeneity analysis, which identifies patterns in the causal relationship between a specific feature and the target variable across different groups, a policy tree guides strategic decisions by highlighting specific conditions under which the presence of a porch is most influential in shaping housing prices. This approach goes beyond understanding variations; it offers a roadmap for targeted approaches in the real estate market.
The policy tree provides a clear set of conditions that stakeholders can use as guidelines for decision-making. Each node in the tree represents a criterion based on property attributes, and the branches depict different outcomes or recommended strategies. The binary Treatment variable in the leaf node streamlines the decision-making process, providing a straightforward recommendation regarding the addition of a porch to the house.
\begin{figure}[h!]  
  \includegraphics[width=\linewidth]{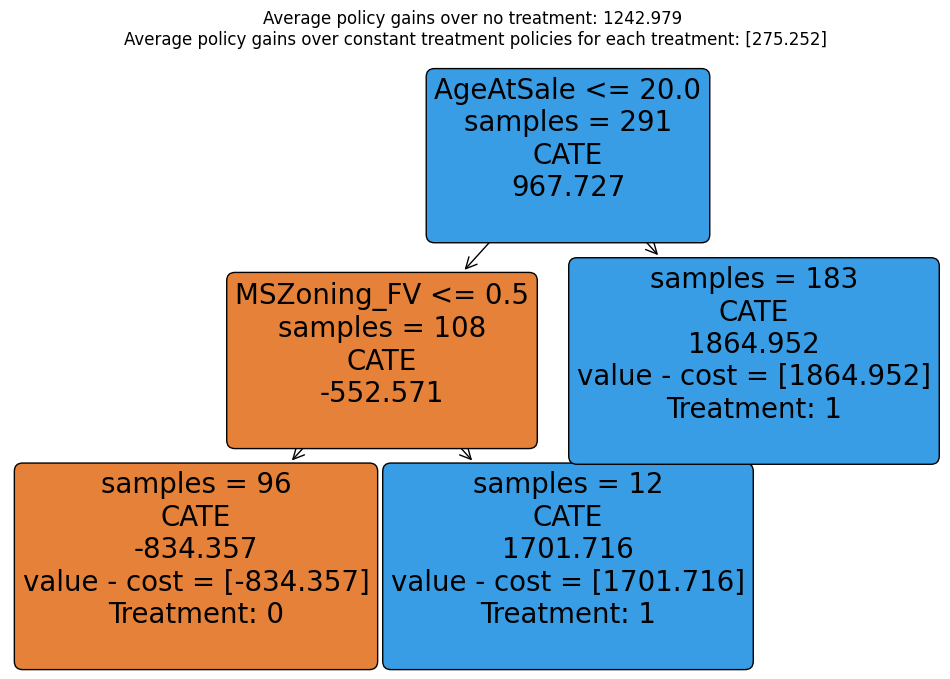}
  \caption{\label{fig:workflow}
    What is the best policy considering cost?}
\end{figure}

The analysis of the policy tree for the 'HasPorch' feature uncovers noteworthy implications for the economic outcomes associated with adding a porch to residential properties. Specifically, houses with an age at sale of less than 20 years exhibit a positive average Conditional Average Treatment Effect (CATE) of 967.727 dollars. This implies that, on average, introducing a porch to these younger houses contributes positively to housing prices. However, a more granular examination reveals a nuanced scenario where younger houses in a specific zoning area may experience a negative CATE of -552.571 dollars. In this context, the policy suggests that, for this particular group, the addition of a porch might lead to a decrease in housing prices. These findings offer actionable insights for decision-makers in the real estate industry, providing guidance on when the implementation of such policies might be economically beneficial and when it could result in potential losses. Such nuanced understanding is crucial for informed decision-making and targeted strategies in the dynamic housing market.

\subsection{What-If Analysis: Assessing the Impact of Adding a Porch}
To delve deeper into the economic implications of adding a porch to residential properties, we conducted a What-If analysis using the CausalAnalysis module. This analysis simulates the hypothetical scenario of introducing porches to houses that initially lack this feature, allowing us to estimate the potential impact on housing prices. The goal is to provide stakeholders in the real estate industry with actionable insights into the financial consequences of such modifications.

For this analysis, we selected houses from the test set that currently do not have a porch (\texttt{HasPorch} = 0). The \texttt{whatif} function was then applied to these houses, considering the addition of a porch by setting the \texttt{HasPorch} variable to 1 while keeping other features constant.

The results of the What-If analysis indicate that the current average housing price on the test set is 146,936.89.dollars Upon simulating the addition of a porch to houses without one, the average housing price is estimated to increase to 149,649.98 dollars on the test set. This suggests a positive impact on housing prices, with the addition of a porch contributing to a higher average valuation.

Interpreting these results, it becomes evident that introducing porches may enhance the overall value of residential properties. The increase in average housing prices implies a potential return on investment associated with adding this feature. However, it's essential for stakeholders to weigh this against the costs involved in constructing porches, ensuring a comprehensive understanding of the economic implications before making decisions. This What-If analysis serves as a valuable tool for decision-makers, offering a glimpse into the potential financial outcomes tied to specific property modifications.

\subsection{Conclusion and Future Work}

This study analyzed the Ames Housing Dataset using CatBoost and LightGBM models, focusing on feature importance and causal relationships in housing price prediction. Key findings include:

\begin{itemize}
    \item High accuracy in price forecasting using both models, with CatBoost achieving a score of 89.91 and LightGBM 86.48 on the test set
    \item A moderate Spearman rank correlation (0.48 for CatBoost and 0.35 for LightGBM) between SHAP-based feature importance and causally significant features
    \item Insights into the causal impact of specific features, such as porches, on housing prices, revealing nuanced effects based on property age and zoning
    \item What-If analysis demonstrating that adding a porch could potentially increase average housing prices from 146,936.89 to 149,649.98 dollars
\end{itemize}

These results highlight the complexity of aligning predictive modeling with causal understanding in real estate valuation. The study underscores the need for integrated approaches that combine predictive power with causal insights, providing actionable information for stakeholders in the real estate industry.

Future research directions include:

\begin{enumerate}
    \item Expanding the dataset to enhance model generalizability and capture broader market trends
    \item Deepening causal analysis on additional features to provide a more comprehensive understanding of housing price determinants
    \item Applying findings to real-world scenarios and policy development, particularly in urban planning and real estate investment strategies
    \item Collaborating with experts across disciplines, including economics and urban studies, for a more holistic understanding of housing market dynamics
    \item Exploring advanced machine learning techniques to further improve the alignment between predictive importance and causal significance
\end{enumerate}

This work lays the foundation for a more comprehensive approach to housing market analysis, with implications for both theoretical research and practical applications in real estate. By bridging the gap between predictive modeling and causal inference, we pave the way for more informed decision-making in property valuation, investment, and policy formulation.

\bibliographystyle{plain}
\bibliography{references}

\begin{table*}
\centering
\caption{Appendix: Details of the Dataset (Part 1)}
\label{data1}
\begin{filecontents*}{features-part1.csv}
Feature Name, Type, Meaning
Order,Discrete,Observation number
PID,Nominal,Parcel identification number  - can be used with city web site for parcel review.
MS SubClass,Nominal,Identifies the type of dwelling involved in the sale.
MS Zoning,Nominal,Identifies the general zoning classification of the sale.
Lot Frontage,Continuous,Linear feet of street connected to property
Lot Area,Continuous,Lot size in square feet
Street,Nominal,Type of road access to property
Alley,Nominal,Type of alley access to property
Lot Shape,Ordinal,General shape of property
Land Contour,Nominal,Flatness of the property
Utilities,Ordinal,Type of utilities available
Lot Config,Nominal,Lot configuration
Land Slope,Ordinal,Slope of property
Neighborhood,Nominal,Physical locations within Ames city limits (map available)
Condition 1,Nominal,Proximity to various conditions
Condition 2,Nominal,Proximity to various conditions (if more than one is present)
Bldg Type,Nominal,Type of dwelling
House Style,Nominal,Style of dwelling
Overall Qual,Ordinal,Rates the overall material and finish of the house
Overall Cond,Ordinal,Rates the overall condition of the house
Year Built,Discrete,Original construction date
Year Remod/Add,Discrete,Remodel date (same as construction date if no remodeling or additions)
Roof Style,Nominal,Type of roof
Roof Matl,Nominal,Roof material
Exterior 1,Nominal,Exterior covering on house
Exterior 2,Nominal,Exterior covering on house (if more than one material)
Mas Vnr Type,Nominal,Masonry veneer type
Mas Vnr Area,Continuous,Masonry veneer area in square feet
Exter Qual,Ordinal,Evaluates the quality of the material on the exterior
Exter Cond,Ordinal,Evaluates the present condition of the material on the exterior
Foundation,Nominal,Type of foundation
Bsmt Qual,Ordinal,Evaluates the height of the basement
Bsmt Cond,Ordinal,Evaluates the general condition of the basement
Bsmt Exposure,Ordinal,Refers to walkout or garden level walls
BsmtFin Type 1,Ordinal,Rating of basement finished area
BsmtFin SF 1,Continuous,Type 1 finished square feet
BsmtFinType 2,Ordinal,Rating of basement finished area (if multiple types)
BsmtFin SF 2,Continuous,Type 2 finished square feet
Bsmt Unf SF,Continuous,Unfinished square feet of basement area
Total Bsmt SF,Continuous,Total square feet of basement area
Heating,Nominal,Type of heating
HeatingQC,Ordinal,Heating quality and condition
Central Air,Nominal,Central air conditioning
\end{filecontents*}
\pgfplotstabletypeset[
    col sep=comma,
    string type,
    every head row/.style={
        before row={\toprule},
        after row={\midrule}
    },
    every last row/.style={after row={\bottomrule}},
    columns/Feature Name/.style={column name=Feature Name, column type=l},
    columns/Type of variable/.style={column name=Type of variable, column type=l},
    columns/Meaning/.style={column name=Meaning, column type=p{9cm}}
]{features-part1.csv}
\end{table*}

\begin{table*}
\centering
\caption{Appendix: Details of the Dataset (Part 2)}
\label{data2}
\begin{filecontents*}{features-part2.csv}
Feature Name, Type, Meaning
Electrical,Ordinal,Electrical system
1st Flr SF,Continuous,First Floor square feet
2nd Flr SF,Continuous,Second floor square feet
Low Qual Fin SF,Continuous,Low quality finished square feet (all floors)
Gr Liv Area,Continuous,Above grade (ground) living area square feet
Bsmt Full Bath,Discrete,Basement full bathrooms
Bsmt Half Bath,Discrete,Basement half bathrooms
Full Bath,Discrete,Full bathrooms above grade
Half Bath,Discrete,Half baths above grade
Bedroom,Discrete,Bedrooms above grade (does NOT include basement bedrooms)
Kitchen,Discrete,Kitchens above grade
KitchenQual,Ordinal,Kitchen quality
TotRmsAbvGrd,Discrete,Total rooms above grade (does not include bathrooms)
Functional,Ordinal,Home functionality (Assume typical unless deductions are warranted)
Fireplaces,Discrete,Number of fireplaces
FireplaceQu,Ordinal,Fireplace quality
Garage Type,Nominal,Garage location
Garage Yr Blt,Discrete,Year garage was built
Garage Finish,Ordinal,Interior finish of the garage
Garage Cars,Discrete,Size of garage in car capacity
Garage Area,Continuous,Size of garage in square feet
Garage Qual,Ordinal,Garage quality
Garage Cond,Ordinal,Garage condition
Paved Drive,Ordinal,Paved driveway
Wood Deck SF,Continuous,Wood deck area in square feet
Open Porch SF,Continuous,Open porch area in square feet
Enclosed Porch,Continuous,Enclosed porch area in square feet
3-Ssn Porch,Continuous,Three season porch area in square feet
Screen Porch,Continuous,Screen porch area in square feet
Pool Area,Continuous,Pool area in square feet
Pool QC,Ordinal,Pool quality
Fence,Ordinal,Fence quality
Misc Feature,Nominal,Miscellaneous feature not covered in other categories
Misc Val,Continuous,Value of miscellaneous feature
Mo Sold,Discrete,Month Sold (MM)
Yr Sold,Discrete,Year Sold (YYYY)
Sale Type,Nominal,Type of sale
Sale Condition,Nominal,Condition of sale
SalePrice,Continuous,Sale price
\end{filecontents*}
\pgfplotstabletypeset[
    col sep=comma,
    string type,
    every head row/.style={
        before row={\toprule},
        after row={\midrule}
    },
    every last row/.style={after row={\bottomrule}},
    columns/Feature Name/.style={column name=Feature Name, column type=l},
    columns/Type of variable/.style={column name=Type of variable, column type=l},
    columns/Meaning/.style={column name=Meaning, column type=p{9cm}}
]{features-part2.csv}
\end{table*}

\begin{table*}
\centering
\caption{Appendix: Casually Insignificant Values from Causal Analysis}
\label{tab:insignificant}
\begin{filecontents*}{insignificant-values.csv}
Feature Name, Value, Reason
MSZoning,C (all),has only 8 instances in the training dataset
Street,Grvl,has only 4 instances
LotShape,IR3,has only 9 instances
Utilities,NoSeWa,has only 1 instances
LotConfig,FR3,has only 3 instances
ExterQual,Fa,has only 9 instances
ExterCond,Po,has only 1 instances
BsmtCond,Po,has only 2 instances
Heating,OthW,has only 1 instances
HeatingQC,Po,has only 1 instances
Electrical,Mix,has only 1 instances
PoolQC,Fa,has only 1 instances
Fence,MnWw,has only 7 instances
MiscFeature,Gar2,has only 1 instances
\end{filecontents*}
\pgfplotstabletypeset[
    col sep=comma,
    string type,
    every head row/.style={
        before row={\toprule},
        after row={\midrule}
    },
    every last row/.style={after row={\bottomrule}},
    columns/Feature Name/.style={column name=Feature Name, column type=l},
    columns/Value/.style={column name=Value, column type=l},
    columns/Reason/.style={column name=Reason, column type=p{9cm}}
]{insignificant-values.csv}
\end{table*}

\end{document}